\documentclass{article} 
\usepackage{iclr2025,times}

\usepackage{amsmath,amsfonts,bm}

\def\eqref#1{equation~\ref{#1}}

\def\1{\bm{1}}

\DeclareMathAlphabet{\mathsfit}{\encodingdefault}{\sfdefault}{m}{sl}
\SetMathAlphabet{\mathsfit}{bold}{\encodingdefault}{\sfdefault}{bx}{n}

\usepackage{hyperref}
\usepackage{graphicx}
\usepackage{url}
\usepackage{multicol}

\usepackage{listings}
\usepackage{xcolor}
\usepackage{makecell}

\usepackage{xspace}

\makeatletter
\DeclareRobustCommand\onedot{\futurelet\@let@token\@onedot}
\def\@onedot{\ifx\@let@token.\else.\null\fi\xspace}

\def\eg{\emph{e.g}\onedot} 
\def\ie{\emph{i.e}\onedot} 
\makeatother

\definecolor{codegreen}{rgb}{0,0.6,0}
\definecolor{codegray}{rgb}{0.5,0.5,0.5}
\definecolor{codepurple}{rgb}{0.58,0,0.82}
\definecolor{backcolour}{rgb}{0.95,0.95,0.92}

\lstdefinestyle{mystyle}{
    backgroundcolor=\color{backcolour},   
    commentstyle=\color{codegreen},
    keywordstyle=\color{magenta},
    numberstyle=\tiny\color{codegray},
    stringstyle=\color{codepurple},
    basicstyle=\ttfamily\footnotesize,
    breakatwhitespace=false,         
    breaklines=true,                 
    captionpos=b,                    
    keepspaces=true,                 
    numbers=left,                    
    numbersep=5pt,                  
    showspaces=false,                
    showstringspaces=false,
    showtabs=false,                  
    tabsize=2
}

\title{ViT Registers and Fractal ViT}

\author{Jason Chuan-Chih Chou \\
Cohere Labs Community \\
Toronto, ON, Canada \\
\texttt{chuanchih@gmail.com} \\
\And
Abhinav Kumar \\
Mathematics and Computing \\
Indian Institute of Technology Roorkee \\
Roorkee, Uttarakhand, India \\
\texttt{abhinav\_k@ma.iitr.ac.in} \\
\AND
Shivank Garg \\
Artificial Intelligence and Data Science \\
Indian Institute of Technology Roorkee \\
Roorkee, Uttarakhand, India \\
\texttt{shivank\_g@mfs.iitr.ac.in} \\
}

\iclrfinalcopy 
\begin{document}

\maketitle

\begin{abstract}
Drawing inspiration from recent findings including surprisingly decent performance of transformers without positional encoding (NoPE) in the domain of language models and how registers (additional throwaway tokens not tied to input) may improve the performance of large vision transformers (ViTs), we invent and test a variant of ViT called fractal ViT that breaks permutation invariance among the tokens by applying an attention mask between the regular tokens and ``summary tokens'' similar to registers, in isolation or in combination with various positional encodings. These models do not improve upon ViT with registers, highlighting the fact that these findings may be scale, domain, or application-specific.
\end{abstract}

\section{Introduction}
Vision Transformer (ViT, \citet{dosovitskiy2020image}) has emerged as a strong alternative to CNNs (convolutional neural networks) for computer vision tasks. Based on the transformer \citep{vaswani2017attention} architecture, it is nearly identical to transformer-based language models (LMs), except that the input tokens are linear projections of pixel patches instead of token embeddings. Similar to encoder LMs, ViT needs to break the permutation invariance of tokens with positional encoding, which has now gone through countless iterations.

While it has always been known that generative LMs based on transformer \textit{decoder}, transformer with causal mask, do not exhibit permutation invariance, it was only reported recently that LMs based on transformer decoder without any positional encoding perform surprisingly well. Known as NoPos \citep{haviv-etal-2022-transformer} or NoPE, it was later shown that in the limit of infinite precision positional info can be fully reconstructed with causal mask as an explanation for its performance \citep{kazemnejad2023the}. We therefore wonder whether similar mask-based positional encoding is possible for ViT. However, preliminary experiments show that applying attention mask to regular tokens destroys performance.

Finally, it was shown recently that a small portion of tokens with a very high norm (outlier tokens) emerge in large ViT after training, which can be mitigated by adding throwaway tokens called registers \citep{darcet2024vision} not tied to the input or contributing to the output. This finding inspires us to test whether it is helpful to add similar tokens not tied to the input and apply attention mask to them to provide positional info, without changing the all-pair attention of regular tokens.

\section{Background and Related Work}
\subsection{Positional Encoding}
There has been many variants of positional encoding for ViT. The original ViT uses learned positional encoding \citep{dosovitskiy2020image}, which may have contributed to its popularity among models including OpenCLIP \citep{ilharco_gabriel_2021_5143773}, DEIT-III \citep{10.1007/978-3-031-20053-3_30}, and DINOv2 \citep{oquab2024dinov}. Experiments reported in \cite{darcet2024vision} that add registers to these three models follow the same practice and use randomly initialized, learned positional encoding for the registers \citep{register-pe}. \cite{9711302} proposes a 2D variant of sinusoidal positional encoding of \cite{vaswani2017attention}, sincos2d, which is found to improve the performance of the ImageNet-1k ViT-S/16 baseline \citep{beyer2022better}. More recently, other positional encodings from the domain of LMs have been ported and tested in ViT, including ALiBi (Attention with Linear Biases, \cite{press2021train}) and RoPE (Rotary Positional Embeddings, \cite{su2021roformer}), which give rise to 2D-ALiBi used in CROMA \citep{fuller2024croma} and RoPE-ViT \citep{heo2025rotary}, respectively.

\subsection{Attention Patterns}

While designed for the purpose of positional encoding, ALiBI can also be considered and implemented as a soft attention mask that reduces attention scores of distant query-key pairs. Conversely, while the causal mask for transformer decoder is originally designed to preserve causality of the output \citep{vaswani2017attention}, it is shown later that decoder LMs perform surprisingly well with just the causal mask and without further positional encoding \citep{haviv-etal-2022-transformer,kazemnejad2023the}. Other modifications of the baseline all-pair attention pattern are usually for the purpose of representation learning, improvement of the training or inference dynamics, or more compute-efficient attention mechanism:

\begin{enumerate}
  \item Representation learning: The practice of adding a special [CLS] token for representation learning goes back to BERT (Bidirectional Encoder Representations from Transformers, \cite{devlin2019bert}). This is followed by the original ViT \citep{dosovitskiy2020image} and vision-language models such as CLIP \citep{radford2021learning}.
  \item Improvement of dynamics: Adding throwaway registers to large ViT to eliminate the artifacts of high-norm outlier tokens \citep{darcet2024vision} falls into this category. The discovery of ``attention sink'': unusually high attention score of the initial token with no semantic relevance and the mitigation of always keeping the initial token key and value for the sliding window of StreamingLLM \citep{xiao2024efficient} can be considered distant parallel in the domain of LMs.
  \item Compute-efficiency: Finally, efficiency of the transformer can be improved through sparsity of the attention mechanism, either with fixed attention pattern or context-dependent token dropping. People have performed extensive experiments with ViT with a variety of attention sparsity \citep{10.5555/3540261.3541789,10204119}, but to our best knowledge the sparse attention pattern has always been applied to the input tokens only instead of additional tokens such as registers.
\end{enumerate}

\section{Methodology}

In fractal ViT, we add ``summary tokens'' which use the same type of positional encoding as that of regular tokens but on smaller grids, in addition to global token [CLS]. Summary tokens are not tied to the input just like registers but we assign $k \times k$ regular tokens to each summary token (``$k$-summary'' in our terminology) depending on the location of the tokens and apply an attention mask to break permutation invariance with a self-similar pattern:

\begin{enumerate}
  \item All $n \times n$ regular tokens attend each other.
  \item All $\frac{n}{k} \times \frac{n}{k}$ summary tokens attend each other, but each summary tokens only attends the $k \times k$ regular tokens assigned to it. The assigned regular tokens also attend back to their shared summary token.
  \item Finally, the global token still attends to all tokens and all tokens still attend back to the global token.
\end{enumerate}

\begin{figure}[t]
    \centering
    \includegraphics[width=\columnwidth]{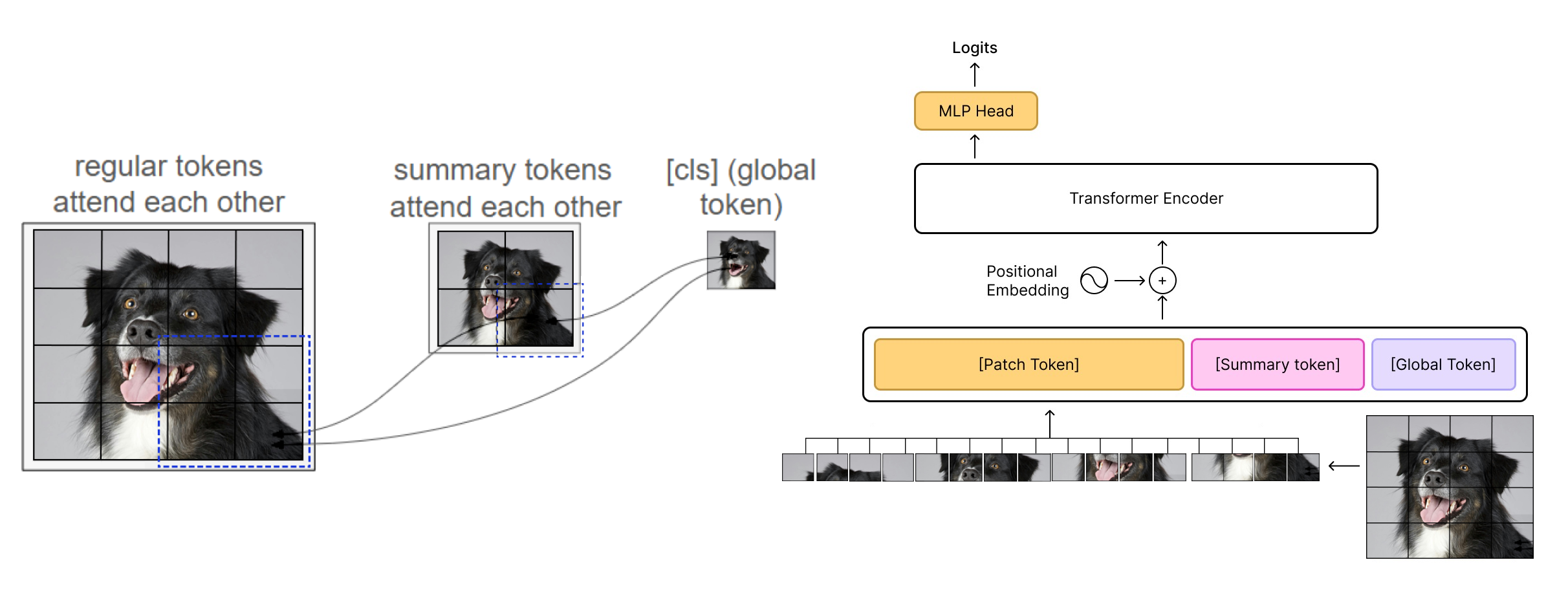} 
    \caption{Fractal ViT architecture. Left: Attention diagram.
    For clarity, only the attention among different types of tokens at the lower-right corner is drawn as arrows. The toy example shown here uses 2-summary that assigns a summary token for every $2\times2$ regular tokens. Right: Regular tokens created from linear projection of RGB values of patches are fed to the transformer encoder along with zero-init summary tokens and global token, optionally after adding the positional encoding.}
    \label{Approach}
\end{figure}

The experiments presented in this paper focus on the simplest case where we have $k^2 \times k^2$ regular tokens, $k \times k$ summary tokens, and one global token, but this pattern can be extended to multiple levels of summary tokens. See Figure \ref{Approach} for a diagram and Appendix \ref{app:mask} for an implementation.

\section{Results}
Here we report top-1 validation set accuracy after training ViT-S/16 for 90 epochs on ImageNet-1k with input resolution 256 (Table \ref{tab:main}). Except the input resolution, positional encoding, and 17 additional tokens, we follow the setup of \citep{beyer2022better}. We test positional encodings \{sincos2d, learned, 2D-ALiBi, none\} (Appendix \ref{app:posenc}), either 17 registers with learned positional encoding \citep{darcet2024vision} or 4-summary with the same type of positional encoding as that of regular tokens but on a $4 \times 4$ instead of $16 \times 16$ grid, and whether fractal mask is applied. In combination with the global token 4-summary results $4\times4 + 1$ additional tokens so the number of tokens is kept constant. sincos2d outperforms both learned positional encoding and 2D-ALiBi, while models without positional encoding perform the worst. Positional encoding of the additional tokens turns out to be inconsequential and fractal mask doesn't improve model performance, even for models without positional encoding.

With fractal mask shown to be unhelpful, we ablate it and further test whether registers or summary tokens are helpful in this setting. Since we are no longer limited to using powers of 4 as the number of regular tokens, we revert to the standard 224 input resolution with the best-performing sincos2d positional encoding (Table \ref{tab:ablation}). Conceptually we either add 59, 17, 9, or 4 registers or add 2-summary, 3-summary, 4-summary, or 5-summary without fractal mask. We follow the rules that

\begin{enumerate}
  \item When $k$ in $k$-summary is small enough we create multiple levels of summary tokens with sincos2d positional encoding but on smaller and smaller grids until doing so results in no more additional tokens.
  \item When $k^m$ doesn't divide the input resolution, we take the floor.
\end{enumerate}

Number of additional tokens is again controlled as we compare ViT with registers to ViT with summary tokens, \eg $\lfloor \frac{14}{2} \rfloor^2 + \lfloor \frac{14}{2^2} \rfloor^2 + \lfloor \frac{14}{2^3} \rfloor^2 = 59$. Differences between summary tokens and registers, \ie positional encoding of the additional tokens, again turn out to be inconsequential. Compared to the baseline of 0 additional tokens, models with 4-59 additional tokens have slightly higher top-1 validation set accuracies at 1-2 standard deviations from the average.

\begin{table}[h!]
\label{tab:main}
\begin{center}
\begin{tabular}{ |c|c|c|c| } 
 \hline
 (pos. enc.) & Register & Summary & \makecell{Summary + Fractal mask\\(Fractal ViT)} \\
 \hline
 sincos2d & \textbf{77.68} & 77.61 & 77.57 \\
 \hline
 learned & \multicolumn{2}{c|}{76.63} & 76.11 \\
 \hline
 2D-ALiBi & --- & 76.16 & 76.26 \\
 \hline
 none & 72.93 & --- & 73.09 \\
 \hline
\end{tabular}
\caption{Top-1 validation set accuracy of ViT-S/16 with 256 input resolution, variants of positional encoding, register or summary tokens, and whether we apply fractal mask for the latter. Since summary tokens use learned positional encoding for (learned, summary tokens) it is identical to (learned, register). We skip (none, summary tokens) since it results in indistinguishable summary tokens while (2D-ALiBi, register) is skipped only because we find it unlikely to yield further insights.}
\end{center}
\end{table}

\begin{table}[h!]
\label{tab:ablation}
\begin{center}
\begin{tabular}{ |c|c|c| } 
 \hline
 (\# of additional tokens) & Register & Summary \\
 \hline
 59 & 77.06 & 77.10\\
 \hline
 17 & 77.08 & 77.01\\
 \hline
 9 & \textbf{77.12} & 77.05\\
 \hline
 4 & 77.07 & 77.02\\
 \hline
 0 (baseline, N=3) & \multicolumn{2}{c|}{76.92$\pm 0.13$}\\
 \hline
\end{tabular}
\caption{Top-1 validation set accuracy of ViT-S/16 with 224 input resolution, different number of additional tokens, and register or summary tokens.}
\end{center}
\end{table}

\section{Conclusion}

Positional encoding of the additional tokens turns out to be inconsequential and fractal mask turns out to be unhelpful. In light of the result, perhaps it is imperative to reexamine the studies that inspired these experiments:

\begin{enumerate}
  \item Transformers without positional encoding (NoPE, \cite{kazemnejad2023the}) that fully relies on the causal mask for breaking permutation invariance and inspired the fractal mask were only tested in the domain of language models. Furthermore, the mask for NoPE applies to tokens directly tied to the input (token embeddings) instead of dedicated attention sink \citep{xiao2024efficient} or additional tokens like registers or summary tokens.
  \item CROMA \citep{fuller2024croma} uses 2D-ALiBi and shows that it outperforms sincos2d in ablation but we do not find it advantageous for the ImageNet-1k ViT-S/16 baseline. Overall CROMA is a distantly-related model for a different purpose (multimodal representation learning for satellite images) so the qestion on what makes the difference remains open. However, perhaps the reasons for 2D-ALiBi's performance in CROMA offered in \citep{fuller2024croma} still hold true:
  \begin{enumerate}
    \item Satellite imagery is rotation-invariant so it is a better match for 2D-ALiBi, which at least exhibits symmetry of dihedral group $D_4$. In contrast, ImageNet is at most symmetric with respect to horizontal flip.
    \item 2D-ALiBi limits attention weights of distant tokens and helps avoid representational collapse due to contrastive objectives, which is not applicable for the ImageNet-1k ViT-S/16 baseline.
  \end{enumerate}
\end{enumerate}

Based on these comparisons, we believe that the following future directions may be worth pursuing:

\begin{enumerate}
  \item Especially in light of the phenomenon of attention sink in Large Language Models (LLMs), it may be worth bringing the idea of fractal masks and summary tokens back to the domain of language models and see if they improve the performance of decoder models or masked encoder models.
  \item Positional encodings may need to be customized to respect the underlying symmetries of the input for better performance. In fact, for the application of satellite imagery, it may be worth trying architectures that fully respect $E(2)$ symmetry such as \citep{pmlr-v216-xu23b} instead of 2D-ALiBi that merely exhibits symmetry of $D_4$.
\end{enumerate}

More recently fractal generative models \citep{li2025fractal} have shown promise in image generation. It is also possible that fractal pattern is more useful in generation than in perception.

\bibliography{iclr2025}
\bibliographystyle{iclr2025}

\appendix
\section{\texorpdfstring{$4$-summary mask snippet}{4-summary mask snippet}} \label{app:mask}

Here is a working function for creating a $4$-summary mask in Python and PyTorch. The implementation we use in production is more general but more complicated.

\begin{lstlisting}[language=Python, label={lst:snippet}, caption=4-summary mask snippet]
def create_fractal_attention_mask(n_h, n_w):
    # Create mask for regular tokens, summary tokens, and global token
    mask_16x16, mask_4x4, mask_global = torch.ones(n_h * n_w, n_h * n_w), torch.ones(n_h * n_w // 16, n_h * n_w // 16), torch.ones(1, 1)
    
    # Combine masks
    mask = torch.block_diag(mask_16x16, mask_4x4, mask_global)

    # Allow 4x4 summary tokens to attend to their corresponding 4x4 regions and vice versa
    for i in range(n_h // 4):
        for j in range(n_w // 4):
            index = n_h * n_w + i * 4 + j
            for row in range(i * 4, i * 4 + 4):
                start = row * n_w + j * 4
                mask[start:start + 4, index] = mask[index, start:start + 4] = 1

    # Allow global token to attend to everything
    mask[-1, :] = mask[:, -1] = 1
    return mask
\end{lstlisting}
\section{Positional Encoding} \label{app:posenc}

Let us define the following notations:
\begin{itemize}
    \item $\mathbf{e}_i \in \mathbb{R}^d$ denotes the patch embedding for position $i$
    \item $d \in \mathbb{N}$ represents the embedding dimension
    \item $\boldsymbol{\pi}_i \in \mathbb{R}^d$ represents the learned position embedding for position $i$
    \item $\mathbf{t}_i \in \mathbb{R}^d$ denotes the final token representation at position $i$
\end{itemize}

The positional encoding mechanism maintains spatial information within the transformer architecture. Traditional Vision Transformers employ learned position embeddings $\boldsymbol{\pi}_i \in \mathbb{R}^d$, which are added to the patch embeddings to form tokens:

\begin{equation}
    \mathbf{t}_i = \mathbf{e}_i + \boldsymbol{\pi}_i
\end{equation}

\subsection{ALiBi: Position-Aware Attention Mechanism}

Let us define:
\begin{itemize}
    \item $\mathbf{q}_i \in \mathbb{R}^d$ denotes the query vector at position $i$
    \item $\mathbf{k}_j \in \mathbb{R}^d$ denotes the key vector at position $j$
    \item $h \in \{0, 1, ..., n - 1\}$ represents the attention head index, assuming that we have $n$ attention heads
    \item $m(h) \in \mathbb{R}^+$ is a head-specific slope parameter
    \item $a_{ij}^h \in \mathbb{R}$ represents the attention score between positions $i$ and $j$ for head $h$
\end{itemize}

The ALiBi mechanism modifies attention computation through position-dependent biases. For attention head $h$, the attention score between positions $i$ and $j$ is defined as:

\begin{equation}
    a_{ij}^h = \frac{\mathbf{q}_i^\top\mathbf{k}_j}{\sqrt{d}} - m(h) \cdot |i-j|
\end{equation}

where $m(h)$ follows the geometric sequence:

\begin{equation}
    m(h) = 2^{-\frac{8(h + 1)}{n}}, \quad h \in \{0, 1, ..., n - 1\}
\end{equation}

\subsubsection{2D ALiBi Extension}

For the 2D variant, we define:
\begin{itemize}
    \item $(i,j), (k,l) \in \mathbb{N}^2$ represent spatial coordinates in the 2D grid
    \item $\mathbf{q}_{(i,j)} \in \mathbb{R}^d$ denotes the query vector at position $(i,j)$
    \item $\mathbf{k}_{(k,l)} \in \mathbb{R}^d$ denotes the key vector at position $(k,l)$
\end{itemize}

The attention score is computed as follows:

\begin{equation}
    a_{(i,j),(k,l)}^h = \frac{\mathbf{q}_{(i,j)}^\top\mathbf{k}_{(k,l)}}{\sqrt{d}} - m(h) \cdot \sqrt{(i-k)^2 + (j-l)^2}
\end{equation}

\subsection{Sinusoidal 2D Positional Embeddings}

Let us define:
\begin{itemize}
    \item $h, w \in \mathbb{N}$ represent the height and width of the feature map
    \item $PE_{(y,x,i)} \in \mathbb{R}$ denotes the positional encoding at spatial position $(y,x)$ for dimension $i$
    \item $\omega_i \in \mathbb{R}^+$ represents the frequency for dimension $i$
    \item temperature $\tau \in \mathbb{R}^+$ is a hyperparameter controlling the frequency spectrum
\end{itemize}

For a feature map of dimensions $h \times w$, the position encoding is computed as:

\begin{equation}
    PE_{(y,x,2i)} = \sin(y\omega_i), \quad PE_{(y,x,2i+1)} = \cos(y\omega_i)
\end{equation}
\begin{equation}
    PE_{(y,x,2i+d/2)} = \sin(x\omega_i), \quad PE_{(y,x,2i+d/2+1)} = \cos(x\omega_i)
\end{equation}

where $\omega_i = \tau^{-4i/d}$ determines the frequency for dimension $i$. 

These formulations ensure unique spatial position encoding while maintaining relative positional relationships across scales.

\end{document}